\documentclass[10pt,twocolumn,letterpaper]{article}

\usepackage{iccv}
\usepackage{times}
\usepackage{epsfig}
\usepackage{graphicx}
\usepackage{amsmath}
\usepackage{amssymb}
\usepackage{authblk}
\usepackage[accsupp]{axessibility}  
\usepackage{subfig}
\usepackage{caption} 
\usepackage[breaklinks=true,bookmarks=false]{hyperref}

\iccvfinalcopy 

\def\iccvPaperID{3881} 

\ificcvfinal\fi

\begin{document}

\title{Stochastic Transformer Networks
with Linear Competing Units: Application to end-to-end SL Translation}



\newcommand*\samethanks[1][\value{footnote}]{\footnotemark[#1]}
\author[1]{Andreas Voskou\thanks{ ai.voskou@edu.cut.ac.cy} }

\author[1]{Konstantinos P. Panousis}
\author[2]{Dimitrios Kosmopoulos}
\author[3]{\\Dimitris N. Metaxas}
\author[1]{Sotirios Chatzis}

{
    \makeatletter
    \renewcommand\AB@affilsepx{: \protect\Affilfont}
    \makeatother

    \makeatletter
    \renewcommand\AB@affilsepx{, \protect\Affilfont}
    \makeatother

\affil[1]{Cyprus University of Technology}
\affil[2]{University of Patras}
\affil[3]{Rutgers University, New Jersey}

}

\maketitle
\ificcvfinal\thispagestyle{empty}\fi

\begin{abstract}
   Automating sign language translation (SLT) is a challenging real-world application. Despite its societal importance, though, research progress in the field remains rather poor. Crucially, existing methods that yield viable performance necessitate the availability of laborious to obtain gloss sequence groundtruth. In this paper, we attenuate this need, by introducing an end-to-end SLT model that does not entail explicit use of glosses; the model only needs text groundtruth. This is in stark contrast to existing end-to-end models that use gloss sequence groundtruth, either in the form of a modality that is recognized at an intermediate model stage, or in the form of a parallel output process, jointly trained with the SLT model. Our  approach constitutes a Transformer network with a novel type of layers that combines: (i) local winner-takes-all (LWTA) layers with stochastic winner sampling, instead of conventional ReLU layers, (ii) stochastic weights with posterior distributions estimated via variational inference, and (iii) a weight compression technique at inference time that exploits estimated posterior variance to perform massive, almost lossless compression. We demonstrate that our approach can reach the currently best reported BLEU-4 score on the PHOENIX 2014T benchmark, but without making use of glosses for  model  training, and with a memory footprint reduced by more than 70\%.
\end{abstract}

\section{Introduction}
\label{sec:intro}

The Sign Languages (SLs) are the native languages of the Deaf and therefore they are the main communication means within the Deaf communities. The SLs are rich visual languages, that convey information through  multiple modalities, which are of complementary nature. Specifically, SLs utilize both manual (hand shape, movement and pose), as well as non-manual modalities (e.g., facial expressions, lip movements, head movements, shoulders and torso), to convey salient meanings \cite{Quer2017}.

Exploiting the latest advances in computer vision and machine learning to facilitate the communication of SL-speakers with SL non-speakers is an endeavor of high potential impact to the livelihoods of the Deaf. Automating the process of converting SL video to written language is the goal of SLT (e.g., \cite{Camgoz18,Camgoz20,CamgozKHB20b,yin2020,Orbay2020,partaourides2020variational}). This has proven to be a hard task for computer vision algorithms, as a natural consequence of the syntax, of the complex entailed  gestures, and of the multitude of concurrent modalities that are combined to convey a unique meaning. 

Due to these challenges, the computer vision community has traditionally focused on recognizing  sequences of sign \emph{glosses}. These are natural language words that attempt to encode the meaning of SL signs, forming a minimal dictionary of indicative lexical items. Thus, the combination of glosses pertaining to some SL video does not constitute translation in natural language; yet, it can help a non-SL speaker get a feeling of what the SL speaker is talking about. The process of pinpointing glosses in SL videos is usually referred to as sign language recognition (SLR). This distinction is important, as the grammar and the structure of sign and spoken languages are very different. These differences are reflected in the outcome of SLR, whereby there is no simple way of associating recognized glosses to actual words/phrases in natural language. This renders SLR outcomes of limited usefulness in real-world applications. 

In an effort to alleviate the limited usefulness of SLR while, at the same time, improving the translation quality of SLT systems, several researchers have recently considered methods that combine SLR with SLT \cite{Camgoz18, Camgoz20, yin2020, CamgozKHB20b}. Specifically, existing methods choose among two alternatives: (i) perform SLR and then translate the sequence  of detected  glosses into natural language (S2G2T); and (ii) train a multitask Deep learning model that jointly performs SLR and SLT, in a way that the representations learned in the intermediate layers  are shared among tasks (S2(G+T)). In the most recent works in the field, this is effected by exploiting a state-of-the-art framework for sequential data modeling, namely the Transformer network \cite{Vaswani2017}.

 Transformer networks \cite{Vaswani2017} currently constitute the state-of-the-art paradigm for sequential data modeling; this includes both sequence-to-sequence modeling tasks and (autoregressive) density modeling tasks. The main principle of Transformer networks, which sets them apart from all previous deep learning approaches for sequential data, consists in the use of a neural attention-based mechanism, dubbed self-attention; this captures (long) temporal dynamics within a modeled sequence. Specifically, self-attention is a dot product attention \cite{luong2015effective} that draws all queries, keys and values from the same sequence. This way, self-attention is the key mechanism that allows for each position within a sequence to attend all the others; this enables capturing long-range dependencies in the data. In addition, it enables high-scale parallelization of computation, which previous approaches (with recurrent connections) cannot afford.

Existing Transformer network formulations are widely founded upon Dense layers with ReLU activation functions. However, several recent works have shown that, by using activation functions employing some sort of stochasticity in their operation, one can yield a considerable performance improvement, especially in hard machine learning tasks. In this context, \cite{panousis2019nonparametric} yielded a considerable performance improvement, without increasing the number of trainable model parameters, by: (i) Replacing ReLU units with blocks of \emph{stochastically competing} local winner-takes-all (LWTA) linear units. Specifically, each layer is split into blocks of \emph{linear} units. At each time, only one of the units within a block passes its activation output to the following layer; that is the winner unit. All the rest are zeroed out, thus passing zero values to the following layer. Winner selection is performed on the basis of a stochastic sampling procedure, whereby the greater the unit activation value the higher the probability of it being sampled as the winner. (ii) Performing an (approximate) Bayesian treatment of the layer parameters (connection weights), whereby the model infers a full variational posterior over layer weights, instead of simple point-estimates.

In this work, we draw inspiration from these advances, seeking an SLT approach that yields significantly improved SL translation accuracy. Our most important goal is to devise an end-to-end SLT modeling approach that completely obviates the need of using SLR groundtruth information (glosses) as part of the model pipeline; that is, either as an intermediate recognition step (S2G2T paradigm), or as a joint task used to facilitate optimization of the learned intermediate input representations (S2(G+T)). Achieving this goal may greatly facilitate progress in the field, since constructing gloss sequences for large training data corpora is an extremely costly and time-consuming process. In addition, our goal is to contribute an SLT method with reduced memory requirements at inference time, as this is important for real-world applications of our technology.

To this end, we devise a novel formulation of Transformer networks, built of Dense layers that comprise the following innovative arguments: (i) LWTA dense layers with stochastic winner sampling, as opposed to conventional ReLU layers; (ii) stochastic connection weights, across the network, with Gaussian posteriors fitted under a variational Bayes rationale; and (iii) a trained network compression scheme, which exploits the estimated variance of the fitted variational posteriors of the layer weights. We employ this novel Transformer network paradigm to formulate an end-to-end SLT model which does not use gloss sequence groundtruth throughout its modeling pipeline. We demonstrate that the proposed method achieves comparable or better results than the state-of-the-art in the most prominent SLT benchmark, namely PHOENIX 2014T. At the same time, our devised model imposes a significantly lower memory footprint compared to the state-of-the-art.

The remainder of this paper is organized as follows: In Section \ref{sec:soa}, we briefly present the recent related work in the field of SLT and SLR, putting more emphasis on the latest advances that make use of Transformer networks. In Section \ref{sec:method}, we present the proposed SLT method; we first introduce our novel modeling rationale; subsequently, we devise appropriate training and inference algorithms; then, we elaborate on the model compression process, which we eventually use to obtain a scalable, end-to-end trainable SLT model. In Section \ref{sec:experimental}, we perform a thorough experimental evaluation of our proposed approach, combined with a deep ablation study. To this end, we use the PHOENIX 2014T dataset. Finally, in Section \ref{sec:conclusion} we conclude this paper, summarizing our results.   

\section{Related Work}
\label{sec:soa}



SLT has been widely treated as a recognition problem (see \cite{Koller2020quant} for a detailed list). Initial approaches sought to recognize individual and well-segmented signs, using discriminative or generative methods under a time-series classification framework; examples include hidden Markov models (HMMs), e.g., \cite{Chat09, Vogler04, Lang2012}, dynamic time warping, e.g., \cite{Alon09, Licht08}, and conditional random fields, e.g., \cite{Yang2006, Yang2010}. These methods involved hand-crafted features; more recently, deep learning methods offered some better representations, such as those stemming from CNNs, e.g., \cite{Pigou2015,Neverova2016}.

This approach to SLT is, however, of limited real-world usefulness, as it yields a set of words with rather incoherent context structure, as opposed to a natural language outcome. Thus, SLT with continuous recognition is a far more realistic framework, but is also much more challenging \cite{Koller2015, Koller2017, Aloysius2020}. The challenge is due to epenthesis (insertion of extra visual cues into signs), co-articulation (the ending of one sign affects the start of the next), and spontaneous sign production (which may include slang, special expressions, etc.). To address the problem, \cite{Koller2020} used a model comprising a CNN-LSTM network to generate features, which are then fed to HMMs that perform inference via a variant of the Viterbi algorithm. In a similar fashion, \cite{Cui2017} used a bi-directional LSTM fed with features from a CNN; \cite{Molchanov2016} used a 3D CNN combined with a penultimate connectionist temporal classification (CTC) layer \cite{Graves2006}. In \cite{ZhouZZL20}, a network dubbed SMTC is proposed, which combines multiple cues from pose and image (hands, face, holistic) in multiple scales, fed to a CTC penultimate layer.

Despite this progress, these works are not capable of scaling to natural language dictionaries of large size. On the contrary, they are typically implemented on either: (i) small dictionaries of relevance to specific real-world scenarios; or (ii) a set of natural language words that attempt to encode the meaning of SL signs in a succinct manner, thus forming a minimal dictionary of indicative lexical items (glosses). Indeed, recognition of glosses is often utilized so as to break the SLT task into two separate tasks of translating signs-to-glosses and, then, glosses-to-text.

These shortcomings have been greatly ameliorated by utilizing Transformer networks. Transformers allow for scaling SLT to real-world natural language dictionaries, while also dramatically increasing the obtained translation performance. This is even more profound when combining SLT with an SLR process, either as an intermediate task, or even in the context of a multitask learning scheme. More specifically, in \cite{Camgoz20}, the authors use a Transformer network to perform translation in an end-to-end fashion. In essence, they propose an S2(G+T) architecture: They postulate a Transformer network to perform S2T; in parallel, they use the encoder part of the Transformer to predict the respective gloss sequence groundtruth. The latter SLR task is performed via a penultimate CTC layer over all possible gloss alignments. Training is performed jointly for the whole structure (both tasks). This way, \cite{Camgoz20} managed to achieve the then highest BLEU-4 score reported on PHOENIX 2014T, equal to 21.80. In addition, the authors also show that using only the end-to-end trainable Transformer network (with no use of gloss sequence groundtruth), they can obtain an SLT BLEU-4 score of 20.17 on PHOENIX 2014T. 

This important breakthrough has spurred fresh research interest in the field, with many recent works building upon and extending this framework. For instance, \cite{CamgozKHB20b} propose to split the visual signal into three different streams: manual, face and body pose. On this basis, they devise a Transformer network with a novel multi-channel attention mechanism, to process the multistream signal. This yielded end-to-end SLT BLEU-4 scores of up to 21.32 on PHOENIX 2014T (without use of glosses). Analogous advances have also been reported on hybrid approaches. For instance, \cite{yin2020} propose an S2G2T hybrid whereby 
Spatial-Temporal Multi-Cue (STMC) networks \cite{ZhouZZL20} are used for gloss recognition; these subsequently feed the recognized gloss sequences to a 2-layered Transformer. This S2G2T network achieves a BLUE-4 score of 24.00; a score of 25.40 is obtained by using an ensemble of such networks. 

At this point, it is important to note that Transformer-based networks which utilize gloss sequence groundtruth currently yield the best reported BLEU-4 scores. The availability of gloss sequences may also be useful for system explainability, but it comes with significant costs: Training in the case of such models entails segmentation/alignment of glosses (via Viterbi decoding, a CTC layer, or similar methods). This, in turns, requires the availability of the possible gloss sequences to be aligned. The alignment process itself incurs additional computations, which are meaningful when addressing SLR, but not necessarily in the case of SLT. Most importantly, the groundtruth of possible gloss sequences is not trivially obtainable; this is especially the case with realistic unconstrained scenarios, which may involve large vocabularies and complex syntax.

\section{Proposed Approach}
\label{sec:method}

\subsection{Conventional Transformer networks}

Before we introduce our proposed approach, we revisit the main principles of Transformer networks. Transformers comprise an encoder module and a decoder module. The encoder is presented with the input sequence, after application of positional encoding (PE), according to the rule

\begin{equation}
\resizebox{0.91\hsize}{!}{%
$\textrm{PE}_{(pos,2i)} = \textrm{sin}(\frac{pos}{10000^{\frac{2i}{d}}}),  \; \textrm{PE}_{(pos,2i+1)} = \textrm{cos}(\frac{pos}{10000^{\frac{2i}{d}}})$%
}
\end{equation}
where $pos$ is a position in the sequence, $i$ is an index, and $d$ the total size of the encoding. Then, it learns to extract a higher-level representation that entails salient temporal dynamics that may unfold over long horizons. To this end, the encoder module is built of a stack of self-attention layers, each of which is paired with two immediately succeeding Dense layers, one with ReLU units and a linear one.

On the other hand, the decoder module is presented with the so-obtained input sequence encoding, and learns to generate the corresponding output sequence. In this context, the decoder module capitalizes upon (possibly multiple) encoder-decoder attention layers; this allows for capturing the salient correlations between input and output sequence patterns in a \emph{continuous} manner. These attention layers are interleaved by preceding, decoder-side, self-attention layers, and succeeding pairs of Dense layers, the former with ReLU units and the latter with linear units. 

In all cases, the attention mechanisms are implemented as the multi-headed variant of dot-product attention. That is, considering a set of keys $K$, queries $Q$, and values $V$, attention computes a linear transformation of the form 
\begin{equation}
\label{eqn:scaledatt}
head = \mathrm{softmax}\big(\frac{KW^kQW^q}{\sqrt{d}}\big)VW^v
\end{equation}
where $d$ is the dimensionality of the input, and $W^\cdot$ are trainable parameter sets. Rule (2) is applied multiple times (as many as the number of heads), with different parameters sets each time. The outcomes are eventually linearly combined to generate the final multihead attention layer output:
\begin{equation}
\label{eqn:multihead}
MultiHead = \mathrm{Concatenate}(head_1,....,head_i)W^m
\end{equation}

\begin{figure*}%
    \centering
    \hspace{-5mm}
     \qquad
    \subfloat{\includegraphics[scale=0.29]{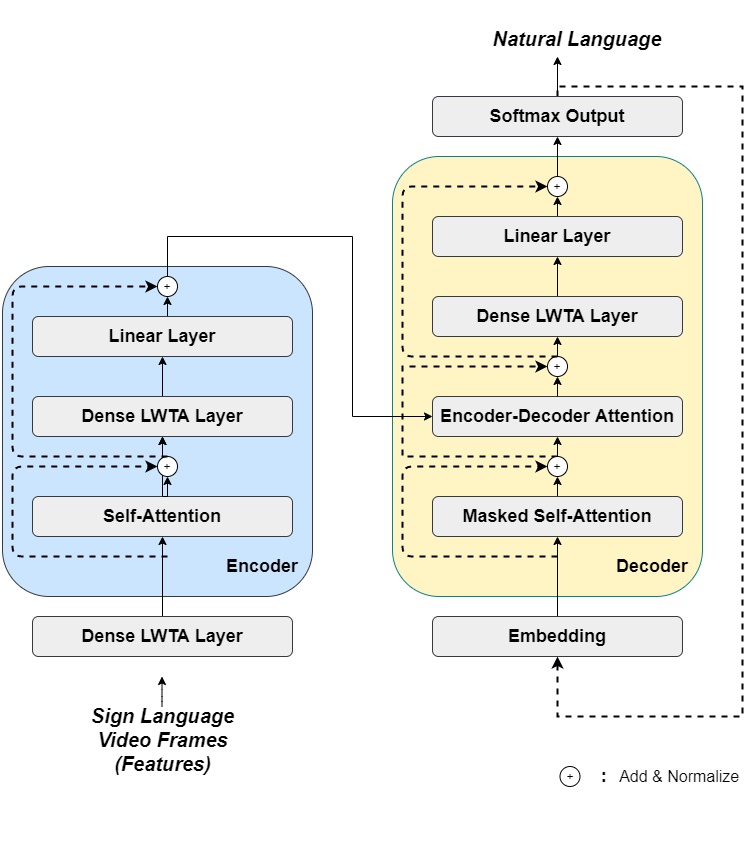} }
   \qquad
   \subfloat{\includegraphics[scale=0.15]{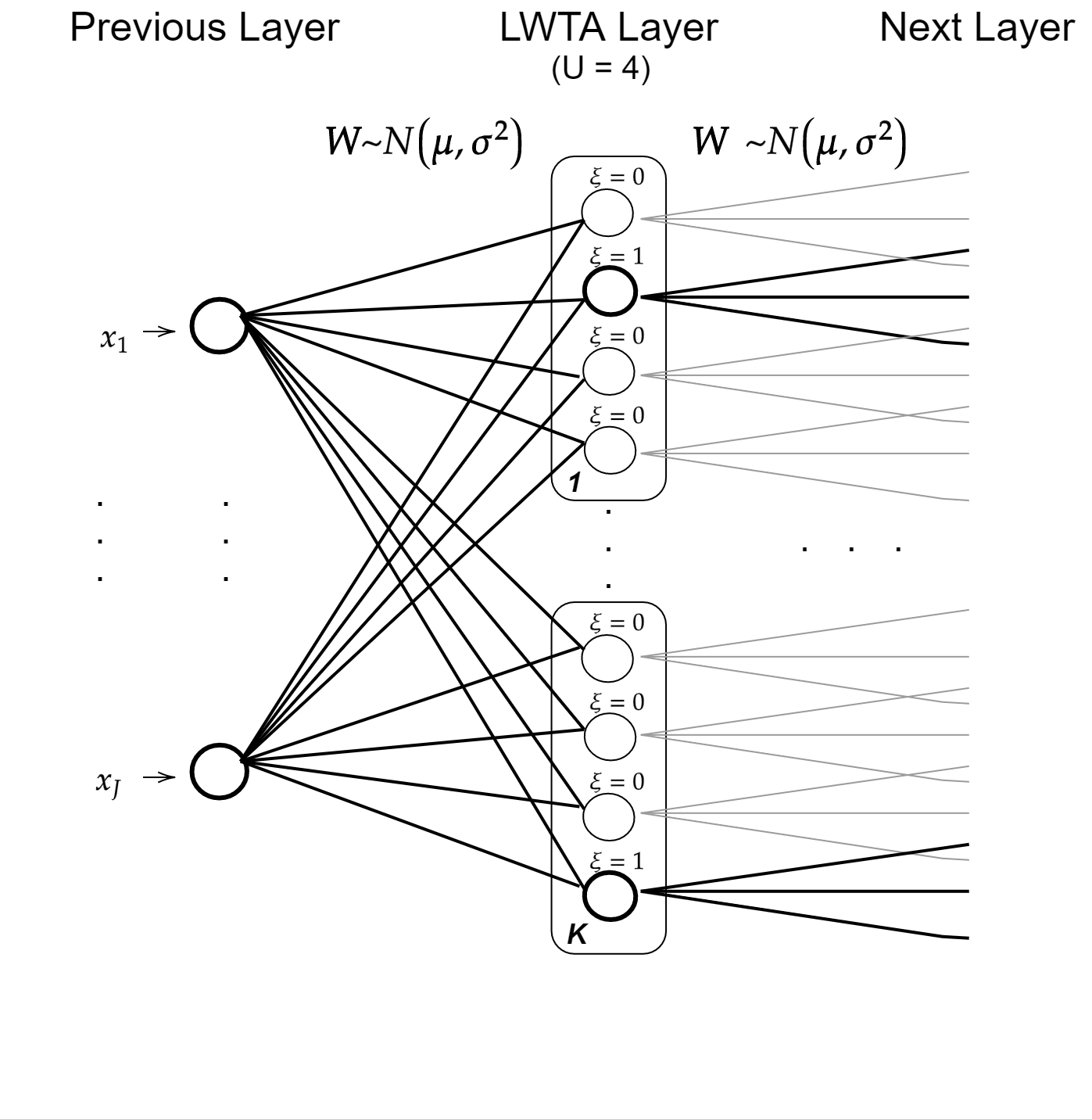} }
    \caption{ Proposed Approach. (a) The Proposed Transformer network for end-to-end SLT. (b) A graphical illustration of the proposed LWTA layers. Rectangles depict LWTA blocks, while circles therein represent competing linear units. The winner units are denoted with bold contours ($\xi=1$). All edges correspond to Gaussian-distributed weights.} %
    \label{fig:fullmodel}%
\end{figure*}

\subsection{Stochastic Transformers with linear competing units}

Let us denote as $\boldsymbol{x}\in\mathbb{R}^{J}$ an input representation vector fed to some dense ReLU layer of a Transformer network, comprising $J$ features. This layer is presented with a linear combination of the inputs, obtained via a weights matrix $\boldsymbol{W}\in\mathbb{R}^{J\times K}$, and produces an output vector $\boldsymbol{y}\in\mathbb{R}^{K}$, which is fed to the subsequent layer. In our approach, this mechanism is replaced by the introduction of LWTA blocks, each containing a set of competing linear units. The layer input is originally presented to each block, via different weights for each unit; thus, the weights of the connections are now organized into a three-dimensional matrix $\boldsymbol{W}\in \mathbb{R}^{J\times K \times U}$, where $K$ denotes the number of blocks and $U$ is the number of competing units therein.

Under our approach, within each block these linear units compute their activations; for the $u$th unit in the $k$th block, we obtain the sum
$\sum_{j=1}^J (w_{j,k,u} ) \cdot x_j$. Then, the block selects one winner unit on the basis of a \emph{competitive random sampling} procedure (described next), and sets the rest to zero. This way, we yield a \emph{sparse} layer output, encoded into the vectors $\boldsymbol{y}\in\mathbb{R}^{K \cdot U}$ that are fed to the next layer. 

In the following, we represent the outcome of local competition between the units in each block via the discrete latent vectors $\boldsymbol{\xi} \in \mathrm{one\_hot(U)}^K$, where $\mathrm{one\_hot(U)}$ is an one-hot vector with $U$ components. These denote the winning unit out of the $U$ competitors in each of the $K$ blocks of a proposed layer, when presented with some input. Using this notation, the output reads
\begin{align}
[\boldsymbol{y}]_{k,u} = [\boldsymbol{\xi}]_{k,u} \sum_{j=1}^J (w_{j,k,u} ) \cdot x_j \; \in \mathbb{R}
\end{align} where we denote as $[\boldsymbol{h}]_l$ the $l$th component of a vector $\boldsymbol{h}$. As we observe, at each time, only one (linear) unit in each LWTA block passes its output to the next layer, while the rest are zeroed out.

Let us now examine the statistical properties of the latent indicator vector $\boldsymbol{\xi}$. To enable data-driven competition between the units within an LWTA block, we postulate that the probability of a unit being sampled as the winner increases with the value of its (linear) output. In other words, we consider sampling from a Discrete posterior to select the winner at each time. On the basis of this rationale, we postulate that, a posteriori, it holds

\begin{equation}
\label{eqn:dense_xi}
q([\boldsymbol{\xi}]_k) =  \mathrm{Discrete}\bigg([\boldsymbol{\xi}]_k \bigg| \mathrm{softmax}\big( \sum_{j=1}^J [w_{j,k,u}]_{u=1}^{U} \cdot x_j \big) \bigg)
\end{equation}

where $[w_{j,k,u}]_{u=1}^{U}$ denotes the vector concatenation of the set $\{w_{j,k,u}\}_{u=1}^{U}$.

On this basis, we obtain a novel variant of Transformer networks, the main operating principles of which  are depicted in Fig. \ref{fig:fullmodel}. We observe that the proposed network entails statistical inference arguments, which bring to the fore stochastic activation principles. Drawing from this inspiration, we proceed to derive a full Bayesian treatment of the obtained network, by also considering that the network parameters themselves are governed by statistical principles. Specifically, we postulate that, throughout the network, all trainable weights are random variables; their (posterior) distributions can be estimated in data-driven fashion. For simplicity, we seek to derive (approximate) independent Gaussian posteriors over the set of trainable weights, $\boldsymbol{w}$:
\begin{equation}
q(\boldsymbol{w}) = \mathcal{N}(\boldsymbol{w}| \boldsymbol{\mu}, \mathrm{diag}(\boldsymbol{\sigma}^2))
\end{equation}
where $\boldsymbol{\mu}$ is the mean and $\boldsymbol{\sigma}^2$ is the variance of the Gaussians.

This concludes the formulation of the proposed Stochastic Transformer networks with competing linear units.

\subsection{Training and inference algorithms}

To train the proposed model, we resort to maximization of the resulting evidence lower-bound (ELBO) of the model. To this end, we need to introduce appropriate prior assumptions regarding the distributions of the winner indicator latent variables, $\boldsymbol{\xi}$ on each LWTA layer, as well as the trainable weights, $\boldsymbol{w}$, throughout the network. For convenience, we postulate a priori spherical Gaussian weights of the form $p(\boldsymbol{w}) = \mathcal{N}(\boldsymbol{0}, \boldsymbol{I})$, and a symmetric Discrete prior over the winners: $[\boldsymbol{\xi}_n]_k \sim \mathrm{Discrete}(1/U)$.

Introducing a mean-field (posterior independence) assumption across layers, we yield the following ELBO:
\begin{align}
\begin{split}
\mathcal{L}(\phi) = \mathbb{E}_{q(\cdot)}\big[\log p(\mathcal{D}|\{\boldsymbol{w, \xi}\}) \big] - \mathrm{KL} \big[\ q(\{ \boldsymbol \xi \}) \ || \ p(\{ \boldsymbol \xi \})\ \big] \\
- \mathrm{KL} \big[\ q(\{ \boldsymbol{w}\}) \ || \ p(\{\boldsymbol{w}\})\ \big]
\end{split}
\label{eqn:elbo}
\end{align}
where $\phi=\{\boldsymbol{\mu}, \boldsymbol{\sigma}^2\}$ is the set of the means and variances of the Gaussian weight posteriors, trained throughout the network in an end-to-end fashion. In this expression, $\mathbb{E}_{q(\cdot)}\big[\log p(\mathcal{D}|\boldsymbol{w, \xi})\big]$ corresponds to the (negative) posterior expectation over the standard categorical cross-entropy error, used for training conventional Transformer networks. All the posterior expectations in the ELBO are computed by drawing Monte-Carlo (MC) samples under (i) the standard reparameterization trick for the postulated Gaussian weights, $\boldsymbol{w}$; and (ii) the Gumbel-Softmax relaxation trick \cite{Maddison2017,jang2017categorical} for the latent winner indicator variables of the LWTA layers, $\boldsymbol{\xi}$. 
On this basis, ELBO maximization is performed using standard off-the-shelf, stochastic gradient techniques; specifically, we adopt Adam \cite{kingma2014adam}. We provide the analytical expression of the ELBO (\ref{eqn:elbo}) in the Supplementary.

Let us now turn to the inference algorithm of our network. At inference time, we \emph{directly draw samples} from the trained posteriors of the \emph{winner selection latent variables}, $\boldsymbol{\xi}$, of the LWTA layers, as well as the trained weight posteriors, $\boldsymbol{w}$, throughout the network. Thus, differently from previous work in the field, the proposed Transformer networks are characterized by a \emph{doubly stochastic} nature, stemming from two different sampling processes. On the one hand, we implement a data-driven random sampling procedure (by sampling from $q(\boldsymbol{\xi})$) to determine the activations of Dense layers in the network (LWTA layers). In addition, we infer the weight values, throughout the network, again based on sampling from the trained posteriors $q(\boldsymbol{w})$\footnote{In detail, inference is performed by sampling the $q(\boldsymbol{\xi})$ and $q(\boldsymbol{w})$ posteriors a total of $S=4$ times, and averaging the corresponding $S=4$ sets of output logits (\emph{Bayesian averaging}).}.

\subsection{A compression scheme}

According to the current standards \cite{sites2008ieee}, computers represent real numbers by a set of bits divided into 3 different subsets:  a single sign bit, a set of $eb$ exponent bits, and a set of $pb$ significant precision bits. Then, the stored value is expressed as a product of three factors:
\begin{equation}
    \mathrm{value} = (-1)^{sign} * 2^{E-2^{eb}-1} * (1+\sum_{i}^{pb}{b_{pb-i}2^{-i}}  )
\end{equation} where $ E= \sum_{i=1}^{eb}{b_i*2^{i-1}} $, and $b_i$ is the $i$th bit. Therein, the second factor determines the maximum and minimum values that can be stored, and the third one determines floating point precision. Typical machine learning implementations (e.g., PyTorch \cite{NEURIPS2019_9015}) employ 8 exponent bits and 23 precision bits (float32 format). Yet, it is now well-established that a variational Bayesian treatment of deep network weights allows for significantly reducing the used bits without damaging  accuracy \cite{panousis2019nonparametric, elan2020}. 

Specifically, the obtained posterior variance of the network weights, $\sigma^2$, constitutes a measure of uncertainty in their sampled values. The higher the associated uncertainty the more the fluctuation of their values. One can leverage this uncertainty information to assess which precision bits, out of the $pb$ available, are significant, and remove the ones which fluctuate too much under approximate posterior sampling. In addition, combining posterior mean, $\mu$, and variance information allows for estimating a confidence interval, that is an interval that sampled weight values may lie within with high probability. Using this information, we can also reduce the number of used exponent bits, $eb$. 

In our work, we perform both these reductions on a layer-wise basis. To this end, we consider the minimum posterior variance $\sigma^2$ of the weights within a layer, as well as the minimum and maximum $\mu$ values.

\subsection{Proposed SLT model}

In our SLT  model, the whole Transformer network is trained from scratch. The input modality is a frame-wise feature sequence, obtained from the whole video frames. These frame-wise features stem from a pretrained Inception network \cite{szegedy2017inception, Koller2020}, in a fashion similar to \cite{Camgoz20}. This input modality is initially fed to an LWTA layer; this yields spatial embeddings that we subsequently feed to the encoder of our proposed Transformer network, illustrated in Fig. \ref{fig:fullmodel}. The output modality, generated from the decoder part of our network, is natural language interpretations. At each time, the decoder is presented with the previous word, which is initially fed to a vanilla linear embedding layer. The whole resulting model is trained in an end-to-end fashion, as described previously. We implement our method considering LWTA blocks of $U=4$ units each.

\section{Experimental Results}
\label{sec:experimental}

In this Section, we perform the comparative assessment of our approach. To this end, we use PHOENIX 2014T dataset \cite{Camgoz18}; this constitutes the most used benchmark in the recent literature. Hence, this benchmark selection allows for optimal and full comparability of our results with the recent related work in the field. The used dataset contains German SL videos of weather forecasts, and corresponding  translations into the German spoken language. They are obtained from 9 different speakers.

\subsection{Experimental setup}

All trained Transformers use embedding sizes of 512 and 8 attention heads. Weight posterior means and variances are initialized by means of Kaiming uniform initialization \cite{he2015delving}. Conventional models, for which we obtain point-estimates (as opposed to weight posteriors), are initialized by employing Xavier normal \cite{glorot2010understanding}. Gumbel-Softmax temperature is set to $T = 1.69$ for training and $T=0.01$ for inference. In all cases, we use Adam \cite{kingma2014adam} with a learning rate of 0.001 ($\beta_{1}=0.9, \beta_{2}=0.998$), and a batch size of 32. During training, we evaluate the networks on the validation set every 80 iterations, and decrease the learning rate by 20\%  if the validation does not improve for 5 consecutive iterations. Training ends when the learning rate falls bellow a minimum of 0.0001. This evaluation process during network training is performed via greedy decoding. At inference time, evaluation on the test set is performed by means of Beam-Search; we perform several runs to determine optimal beam-size in all cases. Our main reference metric for assessing translation quality is the BLEU-4 score. 

Our implementation is developed in Pytorch \cite{NEURIPS2019_9015}, and  based on the "JOEYNMT" \cite{kreutzer-etal-2019-joey},   "Sign language transformer" \cite{Camgoz20}, "Bayesian Compression for Deep Learning" \cite{elan2020}, and "nonparametric Bayesian local-winner-takes-all" \cite{panousis2019nonparametric} frameworks.


\subsection{Benchmarks}
\begin{table}[h]
\caption{State-of-the-art BLEU-4 scores, as of late 2020.}
\begin{center}
\begin{tabular}{|l||c|c|}
\hline
\ Model & Dev & Test  \\ 
\hline\hline
S2T  \cite{Camgoz20}          & 20.69    & 20.17         \\ 
S2(G+T) \cite{Camgoz20}   & 22.12   & 21.80         \\ 
\hline
G2T \cite{Camgoz20}          & 25.35    & 24.54         \\ 
\hline
S2G2T-STMC \cite{yin2020}        & 22.47   &  24.00   \\ 
S2G2T-STMC ensemble \cite{yin2020}   &  24.68   &  25.40     \\ 
\hline
\end{tabular}
\label{tab:benchmark}
\end{center}
\end{table}
Before discussing our results,  we first present some of the latest state-of-the-art methodologies on the considered benchmark, for further reference.  Table \ref{tab:benchmark} summarises the BLEU-4 scores of those models. The first state-of-the-art  model  we  consider  in our  experimental evaluations  is  the  sign-to-text transformer  (S2T)  \cite{Camgoz20}.   Our  SLT  method presented in this paper largely extends upon  this method; thus, we consider this approach as our Baseline.

In addition, we consider three further Transformer-based models, namely  a gloss-to-text (G2T) \cite{Camgoz20}, a sign-to-gloss-and-text (S2(G+T)) \cite{Camgoz20}, and a sign-to-gloss-to-text (S2G2T) \cite{yin2020} model. These methods obtain higher BLEU scores than the basic S2T; the last one actually yields the highest performance reported to-date in the considered benchmark. However, as mentioned in section \ref{sec:soa}, these networks require the possible gloss sequences 
which may be hard to obtain for large training datasets. Specifically, S2(G+T) takes advantage of glosses as a parallel task that facilitates the encoder to obtain better representations; S2G2T utilizes them as an intermediate step, while G2T uses gloss input to obtain natural language (this renders it the least relevant to a real-world SLT task, as it assumes availability of a system that allows for perfect gloss recognition). Additionally, we emphasize that S2G2T employs the computationally burdensome STMC 3-channel recognition network \cite{ZhouZZL20}, while we process the whole frame as a single channel.

\begin{table*}
\begin{center}
\caption{ Proposed Approach: BLEU scores for varying depths.   }
\begin{tabular}{|c||c|c|c|c||c|c|c|c|}

\hline
       Depth     & \multicolumn{4}{c||}{Dev}          & \multicolumn{4}{ c|}{Test}         \\ 

\ \small{encoder-decoder} & BLEU-1 & BLEU-2 & BLEU-3 & BLEU-4 & BLEU-1 & BLEU-2 & BLEU-3 & BLEU-4 \\ 
\hline\hline
1 - 1       &  48.67   &	35.34  &	27.3&	22.03&  47.47&	 34.75&	 26.8&	 21.85    \\ 
2 - 2       &  49.12   &    36.29  &	28.34&	\textbf{ 23.23} & 48.61&	 35.97&	 28.37&	 \textbf{23.65}   \\ 
3 - 3       &  45.68   &    32.87  &    25.72   & 21.66  & 45.84   & 33.40    &25.72 & 21.29    \\ 
\hline
\end{tabular}
\label{tab:exp1}
\end{center}
\end{table*}
\begin{table*}
\begin{center}
\centering
\caption{  Network compression as per Section 3.4: effect on memory requirements and translation quality. }
\begin{tabular}{|c|c|c||c|c||c|c|}
\hline
    Depth   & Average Required & Memory & \multicolumn{2}{c||}{Dev}          & \multicolumn{2}{ c|}{Test}    \\ 

\small{encoder-decoder}& Bits  &  Reduction & BLEU-4 & change & BLEU-4 & change \\ 
\hline\hline
1 - 1     & 9.4    & 70.6\%    & 21.66    & -1.6\%    &22.05  & +0.9\%      \\ 
2 - 2      & 8.8   & 72.3\%    & 23.09    &  -0.6\%    & 23.52   &  -0.5\%    \\ 
3 - 3      & 8.7   & 73.0\%    & 20.82    & -3.8\%     & 20.77   & -2.4\%      \\ 
\hline
\end{tabular}

\label{tab:exp2}
\end{center}
\end{table*}
\subsection{Performance results}
In Table \ref{tab:exp1}, we summarize the performance of our model for network configurations of varying depth. In our setup, an encoder (decoder) of depth $H$ means a module comprising $H$ consecutive submodules of the form depicted on the left (right) hand side of Fig. \ref{fig:fullmodel}(a).

Comparing the best performance reported therein with the summary of state-of-the-art results in Table \ref{tab:benchmark}, we observe that our method outperforms the corresponding S2T baseline approach by 3.48 BLEU-4 scores on the test set. In this context, the best configuration under the proposed modeling approach seems to be the (2-2); this achieves BLEU-4 scores as high as 23.65 on the test set. This performance is superior to the S2(G+T) hybrid network as well, which yields 21.80 BLEU-4 on the test set. This outcome becomes even more prominent if we consider that S2(G+T) imposes much higher computational burden, and most importantly, it requires the possible sequences of glosses as groundtruth.

Subsequently, we examine network compression. By employing the layerwise compression scheme outlined in Section 3.4, we manage to reduce the average required bits for storing network parameters from 32 to less than 10. This fact implies a memory usage of around 30\% that of the baseline SLT Transformer network of \cite{Camgoz20}. In Table \ref{tab:exp2}, we present the average required number of bits throughout the layers of our network. In addition, we show how the compressed network performs in terms of the obtained BLEU-4 scores. These scores are obtained by compressing network parameters, and then re-running inference. Our results show that our compressed network incurs a negligible trade-off in translation accuracy, for massively lower memory needs.

Finally, we turn to the S2G2T ensemble \cite{yin2020}, which still performs better than our approach, yielding a BLEU-4 score of 25.40 (c.f., Table \ref{tab:benchmark}). The key element that renders S2G2T ensembles so potent is the utilization of ensemble decoding. This consists in averaging the predictions of different networks, in order improve the eventual translation quality. Thus, it is worth to examine whether an ensembling scheme can also improve the BLEU-4 scores of our method. To this end, we repeat our experiments with the (2-2)-version, training 10 different network instances with different random seeds. We use the best performing $L=4$ or $L=8$ of the so-obtained 10 networks to perform ensemble-decoding. 

In Table \ref{tab:exp8}, we present the obtained BLEU-4 scores. With $L=8$, our approach yields a BLEU-4 score of 25.59; this is the best BLEU-4 score ever reported in the literature on the considered dataset. We emphasize that we obtain this performance without making use of any predefined gloss sequences that need alignment in the Transformer network pipeline, contrary to \cite{yin2020}. Then, we repeat our ensemble-decoding experiment using the technique of Section 3.4 to perform parameter compression. We obtain a memory footprint reduction similar to the second line of Table \ref{tab:exp2}. As we show in Table \ref{tab:exp8}, for a memory footprint reduced by approximately 70\%, our method remains competitive with \cite{yin2020}.

\subsection{Ablation study}

\subsubsection{How ReLU activations would perform?}

We now scrutinize the proposed stochastic competition-based activation functions. Specifically, we re-implement our method using ReLU and other popular  activation functions  in place of the proposed LWTA layers. We continue, though, to perform a full variational Bayesian treatment of the model, by inferring Gaussian weight posteriors. Since the (2-2)-version of the proposed network was shown to be the most accurate, all the following experiments focus on this configuration.

\begin{table}[h]
\begin{center}
\caption{BLEU-4 scores with Ensemble-Decoding.}
\begin{tabular}{|c||c|c||c|c|}
\hline
          & \multicolumn{2}{c||}{32 bit }          & \multicolumn{2}{ c|}{Reduced }         \\ 

\ $L$ & Dev & Test  & Dev & Test  \\ 
\hline\hline

4      & 24.02    & 24.84    & 24.23    & 24.52       \\ 
8   &  24.88    &\textbf{ 25.59}    & 24.52    & 25.33       \\ 

\hline
\end{tabular}
\label{tab:exp8}
\end{center}

\end{table}

Table \ref{tab:exp4} illustrates the so-obtained experimental outcomes. It is clear that the proposed LWTA activations with 4 units in each block constitute the approach with the best overall performance; in particular, it yields an advantage of more than 1 BLEU-4 units over the commonly used ReLU and the other conventional activation functions.
Further, we also examine how our approach performs if we use a different number of competing units ($U$) per block. 
Table  \ref{tab:exp4} makes apparent that for both $U=2$ and $U=8$  LWTA still yields better scores than ReLU, but larger blocks seem to decrease the performance.
Finally, we perform network compression following the rationale of Section 3.4, and repeat our experiments. As shown in Table \ref{tab:exp4}, ReLU continues to yield approximately 1.0 BLEU-4 units less than our approach (with $U=4$); this corroborates the superiority of the proposed activations.

\subsubsection{Does the variational Bayesian treatment of network weights contribute to SLT accuracy?}

Conversely to the experiments of the previous Section, it is also important to examine whether training full variational posteriors over the network weights does actually offer tangible gains in terms of translation accuracy. To this end, we re-implement our method, making full utilization of the proposed stochastic LWTA activations, but obtaining conventional point-estimates over the network weights. Thus, the set of network weights, $\boldsymbol{w}$, becomes now a parameters set that we optimize during training. Specifically, network training now reduces to maximization of the following ELBO expression:
\begin{align}
\label{eqn:elbo_full}
\mathcal{L}(\boldsymbol{w}) = \mathbb{E}_{q(\boldsymbol{\xi})}\big[\log p(\mathcal{D}|\{\boldsymbol{\xi}\}) \big] - \mathrm{KL} \big[\ q(\{ \boldsymbol \xi \}) \ || \ p(\{ \boldsymbol \xi \})\ \big]
\end{align}
In Table \ref{tab:exp5}, we provide our results, again considering the (2-2)-version of our method, which constitutes its best-performing configuration. Our findings show that, even with point-estimates, we manage to score 2 BLEU-4 units above the S2T Baseline. This outcome is clearly inferior to our full-fledged model. Therefore, we deduce that the variational Bayesian treatment of connection weights, throughout the proposed network, offers important SLT accuracy gains. This comes in addition to allowing for massive memory savings, by following the rationale of Subection 3.4.

\begin{table}
\caption{Activation function comparison (BLEU-4 scores).}
\begin{center}

\begin{tabular}{|c||c|c||c|c|}
\hline
        & \multicolumn{2}{c||}{32 bit }          & \multicolumn{2}{ c|}{Reduced }         \\ 

\ Activation    & Dev & Test  & Dev & Test  \\ 
\hline\hline
ReLU         & 22.42    & 22.61    & 22.17    & 22.67        \\ 
Elu           & 22.63      & 22.56    & 22.19    & 22.32        \\ 
SiLU         & 22.73     & 22.33    & 22.23    & 21.99        \\ 
\hline
LWTA - $U=2$      & 22.99    & 22.82    & 23.12    & 22.37        \\    
LWTA - $U=4$    & 23.23    &\textbf{ 23.65}    & 23.09    &23.52      \\ 
LWTA - $U=8$         & 22.28   & 22.96    & 22.35    & 22.72        \\ 
LWTA - $U=16$         & 22.32   & 22.52    & 22.00    & 22.34        \\ 
\hline
\end{tabular}
\label{tab:exp4}
\end{center}
\end{table}

\begin{table}
\caption{Comparison of variational Gaussian weights to point-estimates (BLEU-4 scores).}
\begin{center}
\begin{tabular}{|c||c|c||c|c|}
\hline
   Weights     & \multicolumn{2}{c||}{32 bit }          & \multicolumn{2}{ c|}{Reduced }  \\ 
\ Type    & Dev & Test  & Dev & Test  \\ 
\hline\hline
\shortstack{Point-Estimates   } & 22.54    & 22.34    & -   & -     \\ 
\shortstack{Variational Gaussian}           & 23.23    & 23.65    & 23.09    & 23.52 \\ 
\hline
\end{tabular}

\label{tab:exp5}
\end{center}
\end{table}

\subsection{ Qualitative  investigation  }
From a qualitative perspective, our translations seem to be of acceptable quality (Table \ref{tab:exp9}). There is a small number of syntactic and grammar errors; most of them are about locations and dates. Moreover, while in many cases the predicted sentence is syntactically different from the 
groundtruth, the resulting meaning remains similar. C.f. the Supplementary for more examples and English translations.  
\begin{table}[h]
\footnotesize
\begin{center}
\caption{Reference (R), single model (S), and ensemble (E).}
\begin{tabular}{|l|}
\hline
\textbf{R: } im süden schwacher wind\\
\textbf{S: } der wind weht meist nur schwach\\
\textbf{E: } der wind weht im süden schwach bis mäßig\\
\hline
\textbf{R: }  am freitag insgesamt viele wolken die regen bringen\\
\textbf{S: } am donnerstag viele wolken hier und da schauer\\
\textbf{E: } am freitag gibt es viele wolken und gebietsweise schauer\\
\hline
\textbf{R: }ganz ähnliche temperaturen wie heute zwischen sechs und elf grad\\ 
\textbf{S: } am bodensee heute nacht nur sechs bis elf grad\\
\textbf{E: }ähnliches wetter heute nacht\\

\hline
\textbf{R: } im westen und nordwesten fallen einzelne schauer . \\
\textbf{S: } im westen und nordwesten gibt es im westen hier und da schauer . \\
\textbf{E: } im westen und nordwesten gibt es im westen einige schauer .\\

\hline 
\end{tabular}

\label{tab:exp9}
\end{center}

\end{table}

\section{Conclusions}
\label{sec:conclusion}

We proposed an SLT method with the following advantages:
(i) no requirement of glossing sequences for training; (ii) state-of-the-art BLEU-4 score on PHOENIX 2014T, competing with methods that require possible gloss sequences and/or multiple streams; and (iii) at least 70\% less memory requirements than the state-of-the-art. We achieved this by devising a Transformer network that: (i) replaces ReLU layers with stochastically competing linear units; and (ii) performs variational Bayesian inference over all connection weights, throughout the network. 

\section*{Acknowledgement}

This research was partially supported by the Research Promotion Foundation of Cyprus, through the grant: INTERNATIONAL/USA/0118/0037 (Dimitrios Kosmopoulos, Dimitris Metaxas), and the  European Union’s Horizon 2020 research and innovation program, under grant agreement No 872139, project aiD (Andreas Voskou, Sotirios Chatzis)

\small
\bibliographystyle{ieee_fullname}
\bibliography{egpaper_final}

\end{document}